%% file: acl_latex.tex
\documentclass[11pt]{article}

\usepackage[preprint]{acl}

\usepackage{times}
\usepackage{latexsym}

\usepackage[T1]{fontenc}

\usepackage[utf8]{inputenc}

\usepackage{microtype}

\usepackage{inconsolata}

\usepackage{graphicx}

\usepackage{colortbl}  
\usepackage{xcolor}  
\usepackage{amsmath}
\usepackage[utf8]{inputenc}
\usepackage[most]{tcolorbox}
\usepackage{pgfplots}
\usepackage{enumitem}
\usepackage[table]{xcolor}
\usepackage{latexsym}

\usepackage{pifont}
\usepackage{booktabs}
\usepackage{amssymb}
\usepackage{adjustbox}
\usepackage{multirow}
\usepackage{caption}
\usepackage{subcaption}
\newcommand{\passany}[1]{\ensuremath{\text{Pass@}#1}}
\newcommand{\passall}[1]{\ensuremath{\text{Pass\textasciicircum}#1}}
\usepackage{float}
\definecolor{nBlue}{RGB}{0,165,249}
\definecolor{nGreen}{rgb}{0, 0.5, 0.2}
\definecolor{nRed}{rgb}{0.8, 0.1, 0.2}
\definecolor{mGreen}{rgb}{0.3, 0.65, 0.4}
\usepackage[ruled, vlined, noend, linesnumbered]{algorithm2e}

\title{\textsc{MaD-RSI}: Model-as-Data for Recursive Self-Improvement of \\ Agent Harnesses}

\author{
\textbf{Yu Cui\textsuperscript{1}} \quad
\textbf{Hong He\textsuperscript{2}} \quad
\textbf{Ruiqing Yue\textsuperscript{3,4}} \quad
\textbf{Sicheng Pan\textsuperscript{1}} \quad
\textbf{Zhuoyu Sun\textsuperscript{1}}  \quad
\textbf{Yifei Liu\textsuperscript{1}}   \\
\textbf{Ting Li\textsuperscript{5}} \quad
\textbf{Wenzhuo Zhu\textsuperscript{5}} \quad 
\textbf{Baohan Huang\textsuperscript{1}} \quad 
\textbf{Haibin Zhang\textsuperscript{6,7}} \quad
\textbf{Cong Zuo\textsuperscript{1}} \quad
\\ 
\textsuperscript{1}Beijing Institute of Technology \quad
\textsuperscript{2}Tsinghua University \\
\textsuperscript{3}Chengdu Institute of Computer Applications, Chinese Academy of Sciences\\ 
\textsuperscript{4}University of Chinese Academy of Sciences \quad
\textsuperscript{5}Beijing University of Technology \\
\textsuperscript{6}Yangtze Delta Region Institute of Tsinghua University, Zhejiang \\
\textsuperscript{7}Jiaxing Key Laboratory of Artificial Intelligence and Cyber Resilience\\
\texttt{cuiyu@bit.edu.cn, bchainzhang@aliyun.com} 
}

\pgfplotsset{compat=1.18}
\begin{document}
\maketitle
\begin{abstract}
Recursive self-improvement (RSI) of agent harnesses has emerged as an effective approach to enhancing the capabilities of large language model (LLM) agents. By searching the harness space for a specific LLM on a given dataset, RSI can discover a harness tailored to the capabilities and behavioral characteristics of that model. However, such optimization often creates a strong coupling between the learned harness and its target LLM, limiting generalization across models and requiring substantial computational resources to adapt to new LLMs. Moreover, existing RSI methods typically optimize a harness using feedback from a single model and evaluate its generalization to other models only after training. This paradigm treats model differences primarily as a criterion for evaluating generalization rather than as valuable evidence for optimization. In this paper, we introduce a new perspective that treats models as data and places the harness at the center of optimization. Building on this perspective, we propose \textsc{MaD-RSI}, a framework that directly optimizes a shared harness across multiple heterogeneous LLMs by incorporating their diverse failure trajectories into the training process. Rather than assessing model differences only after optimization, \textsc{MaD-RSI} leverages these differences as training evidence to guide the self-improvement of a shared harness. Furthermore, we introduce a new evaluation framework with two complementary metrics designed to assess the effectiveness of harness optimization across models. Experiments across multiple datasets and LLMs demonstrate that \textsc{MaD-RSI} improves generalization over existing approaches while reducing training costs.
\end{abstract}

\section{Introduction}

\begin{figure*}[t]
    \centering
\includegraphics[width=0.9\linewidth]{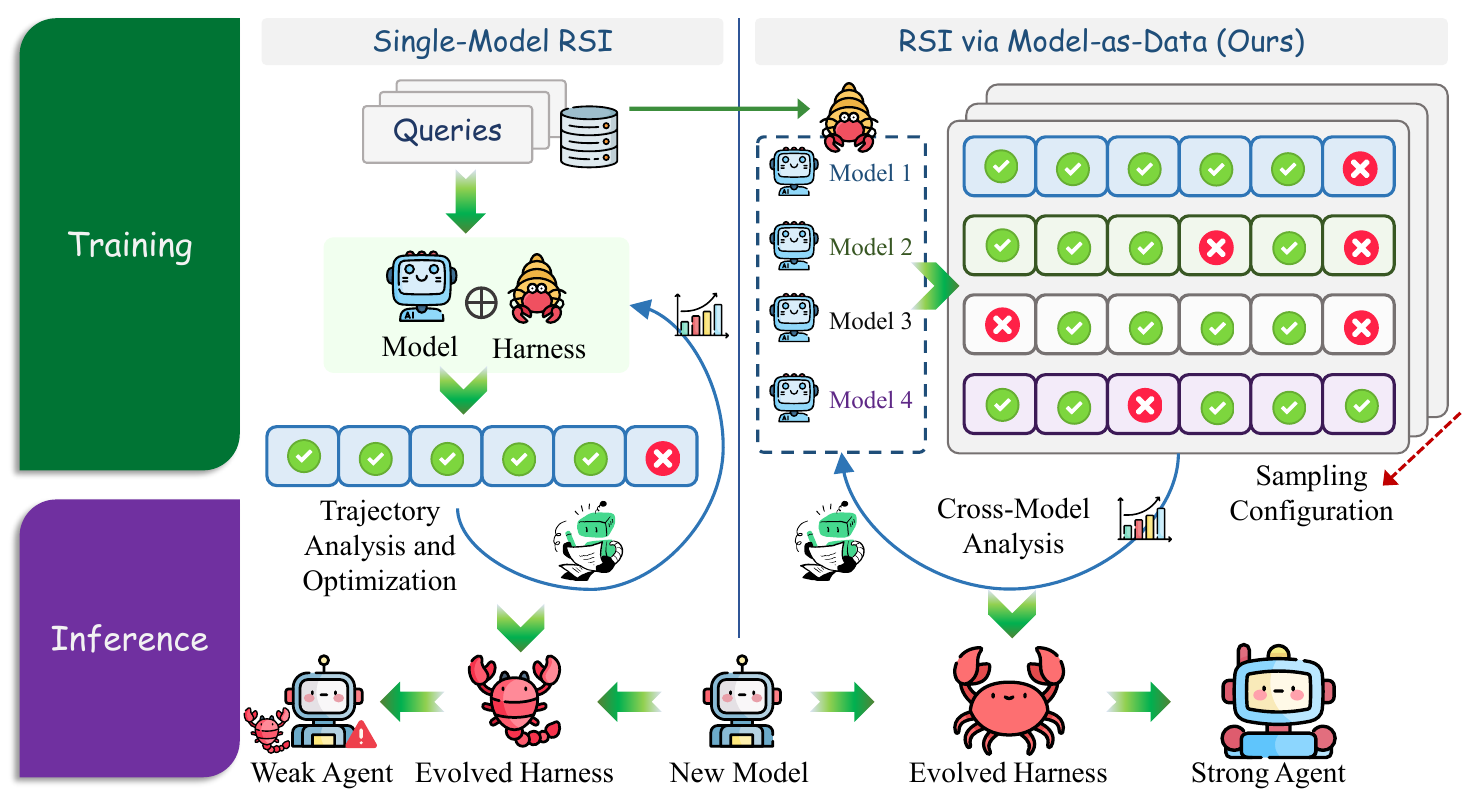}
    \caption{Comparison between \textsc{MaD-RSI} and existing RSI approaches that evolve agent harnesses using feedback from a single backbone model.}
    \label{fig:overview}   
    \vspace{-15pt}
\end{figure*}

Large language model (LLM) agents have achieved strong performance on complex tasks, including reasoning and code generation \citep{yang2024sweagent, yao2023react, chen-etal-2025-locagent}. Their effectiveness depends not only on the underlying LLM but also on the harness. Recent work on recursive self-improvement (RSI) has shown that iteratively searching and refining this harness can enhance agent capabilities without updating model parameters \citep{chan2026auto, duan2026aibuilt, lin2026agentic}. However, many RSI methods optimize a harness using execution feedback from a single target LLM \citep{liu2026solpi}. Such specialization can entangle the learned harness with the capabilities and behavioral characteristics of that model, limiting its portability to other LLMs \citep{qu2026she}. Adapting the harness to a new model may therefore require another costly optimization run, making it difficult to build an ecosystem of reusable and continuously improving harnesses.

This limitation also shapes how RSI is evaluated. A common protocol optimizes a harness with one LLM and assesses its transfer to other models \citep{lee2026metaharness, xia2026rrsi}. In this paradigm, differences among models primarily serve to measure generalization, rather than to guide improvement. Yet heterogeneous LLMs can exhibit different outcomes and execution trajectories on the same task under the same harness. A failure from one model, when contrasted with a success from another, may reveal deficiencies that are difficult to identify from either trajectory alone. These observations suggest a different question: \emph{Can behavioral differences among LLMs serve as training evidence for learning a harness that generalizes across models?}

We address this question through \textit{Models-as-Data}, a perspective that treats heterogeneous LLMs, observed under controlled inference configurations \citep{deepinfer2026} (such as sampling), as sources of behavioral evidence while placing the shared harness at the center of optimization \citep{mehri-etal-2022-lad}. Instead of viewing models solely as separate optimization targets or subjects of subsequent transfer evaluation, we use their successful and failed executions as complementary signals for harness improvement. The goal is to optimize a single harness that can be used with different LLMs independently, not to ensemble multiple models at inference time.

Building on this perspective, we propose \textsc{MaD-RSI}, a framework for RSI of a shared harness using evidence from multiple heterogeneous LLMs. \textsc{MaD-RSI} first organizes responses to shared queries across models and inference configurations into a structured tensor, preserving the conditions under which each outcome is observed. It then selects informative queries using both disagreement among models and coverage of distinct response patterns. Finally, it jointly diagnoses aligned successful and failed trajectories to inform subsequent harness revisions. In this way, behavioral differences across models become inputs to the optimization process rather than observations reserved for evaluation.

To assess RSI methods designed for multiple models, we further introduce an evaluation framework with two complementary measures of optimization effectiveness: \emph{success coverage}, which captures whether at least one model succeeds on a task, and \emph{joint success}, which captures whether all evaluated models succeed on that task. These population measures complement evaluations of transfer to previously unseen LLMs. We study \textsc{MaD-RSI} on legal reasoning, code generation, medical reasoning, and agent safety tasks across heterogeneous LLMs. The results show improvements in performance across models and transfer in some settings, with tradeoffs that vary by task. On the general tasks, optimizing one shared harness also reduces cumulative training cost relative to running separate optimization processes for each development model. More broadly, this study invites a reconsideration of the role of LLMs in RSI: beyond being targets of harness optimization, they can provide the evidence that drives it. We summarize our contributions as follows:  
\begin{itemize}[left=0pt, itemsep=0pt]
    \item \textbf{Models-as-Data perspective.} We formulate an RSI paradigm across models in which behavioral differences among heterogeneous LLMs become evidence for optimizing a shared harness, rather than serving only as a means of evaluating transfer after training.
    \item \textbf{\textsc{MaD-RSI} framework.} We develop a shared harness optimization framework that combines structured response representations, evidence selection based on disagreement and behavioral coverage, and joint diagnosis of successful and failed trajectories.
    \item \textbf{Evaluation and empirical study.} We introduce two complementary population measures for assessing harness optimization across models and evaluate generalization and training cost across diverse tasks and LLMs.
\end{itemize}

\input{latex/Pre}

\input{latex/Methodology}

\input{latex/Experiment}

\input{latex/Related_Work}

\section{Conclusion}
In this paper, we introduce \textit{Models-as-Data}, a new perspective that treats heterogeneous LLMs as behavioral data sources for shared harness optimization. Building on this perspective, we propose \textsc{MaD-RSI}, which leverages behavioral differences across models to guide RSI. We further introduce two complementary metrics to evaluate optimization effectiveness across models. Experiments across diverse tasks and LLMs demonstrate improved generalization in several settings and reduced cumulative training costs compared with independent optimization on general tasks. We hope this work encourages a broader shift in RSI from optimizing harnesses for individual models toward learning reusable harnesses from diverse model behaviors.

\section{Limitations}
This work provides preliminary evidence for the feasibility and potential value of the Models-as-Data perspective, while more comprehensive experimental validation is left for future work. This work is still in progress.

\section{Ethical Considerations}
We used AI assistants for language polishing.


\bibliography{acl_latex}
\onecolumn
\newpage

\appendix

\end{document}

%% file: latex/Pre.tex
\section{Preliminary Analysis}
\label{sec:formulation}
\paragraph{Research Objective.}
The best frozen backbone model for supplying optimization feedback may not be known in advance. Our goal is to improve a single shared harness using evidence from multiple frozen backbone models while limiting total development cost. Let $H$ be a harness, $\mathcal M=\{m_1,\ldots,m_M\}$ a model group, and $\mathcal Q$ a pool with $N$ queries. A sampling configuration $s\in\mathcal S$ specifies decoding controls, with $L=|\mathcal S|$\footnote{In this paper, we use temperature settings as a representative case of inference configurations for our study.}. Execution induces
\begin{equation}
 (\tau,y,c)\sim P_m^H(\cdot\mid q,s),
\end{equation}
where $\tau$ is a trajectory, $y$ an evaluated outcome, and $c$ its cost. Model-as-data treats these conditional behavioral distributions as evidence for harness improvement.

\paragraph{Evaluation.}
Let $y_{irm}\in\{0,1\}$ be the pass event for evaluation query $i$, repetition $r$, and model $m$. With $N_e$ queries and $R_e$ repetitions, define
\begin{align}
 \passany{M}&=\frac{1}{N_eR_e}\sum_{i,r}\max_m y_{irm},\label{eq:any}\\
 \passall{M}&=\frac{1}{N_eR_e}\sum_{i,r}\prod_m y_{irm}.\label{eq:all}
\end{align}
These measure oracle coverage and joint success across models. Both average over all aligned query and trial positions rather than maximizing over repetitions. Trial indices align records but do not imply shared random seeds. Here $M$ counts models, unlike the sample count in conventional $\passany{k}$ evaluation \citep{chen2021eval}.

\paragraph{Cost.}
Shared optimization need not match the best independent run to offer a useful cost tradeoff. Let $H_j$ be a harness optimized with model $m_j$, at training cost $C_j$. Let $H_{\mathrm{mad}}$ be the shared harness, with cost $C_{\mathrm{mad}}$. Fix a higher-is-better population score $S$, such as $\passall{M}$, under one evaluation protocol. Define $S_{\mathrm{best}}=\max_j S(H_j)$ and $C_{\Sigma}=\sum_j C_j$. Under a utility $S-\lambda C$, shared optimization is weakly preferred exactly when
\begin{equation}
 S_{\mathrm{best}}-S(H_{\mathrm{mad}})
 \leq \lambda(C_{\Sigma}-C_{\mathrm{mad}}).
 \label{eq:portfolio-tradeoff}
\end{equation}
Here $\lambda\geq0$ is a cost preference fixed before comparing outcomes, in score units per token. The comparator pays for all independent runs before selecting one harness. This decision criterion allows a score deficit when token savings compensate under the chosen preference.

%% file: latex/Methodology.tex
\begin{figure*}[t]
    \centering
    \begin{subfigure}{0.32\textwidth}
        \centering
        \includegraphics[width=\linewidth]{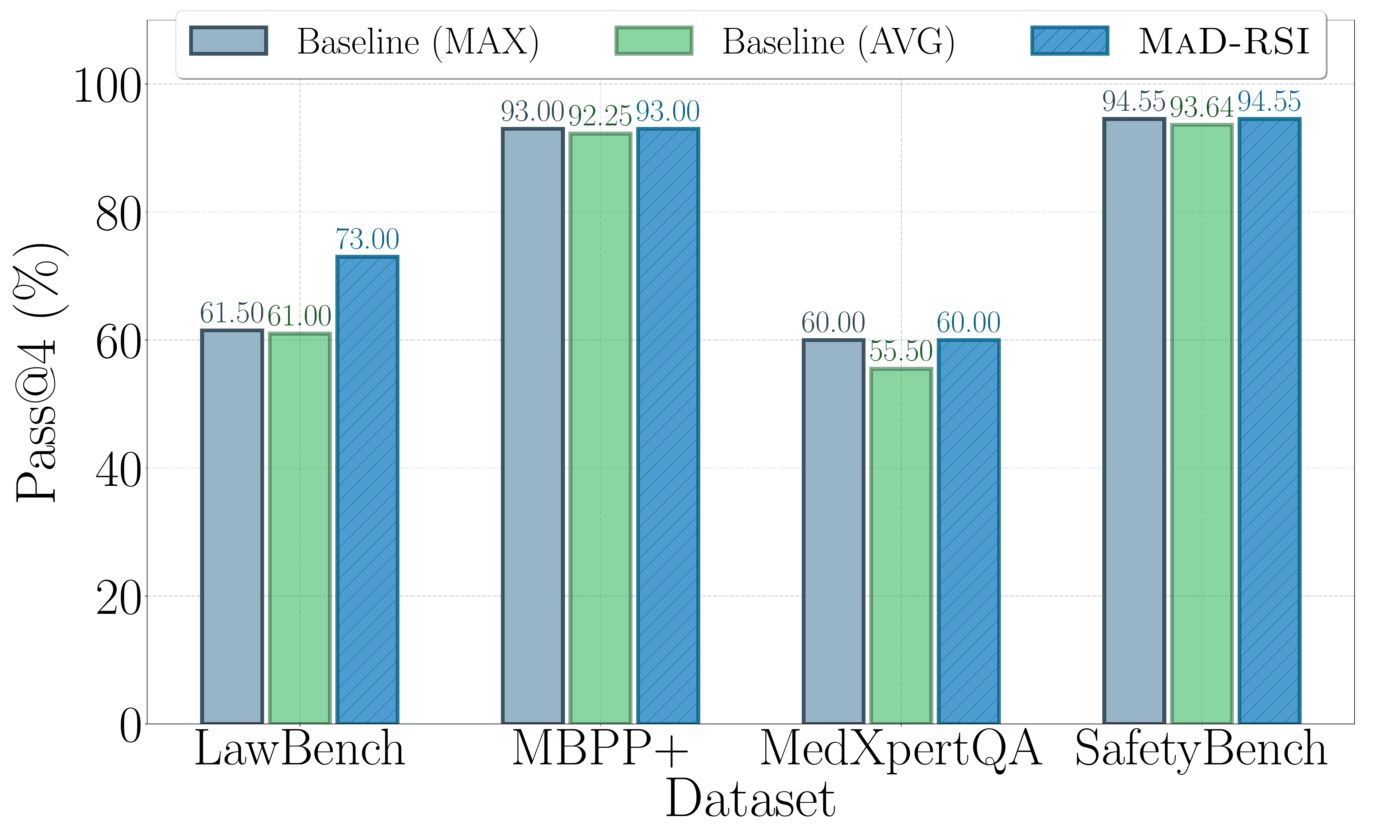}
        \caption{$\passany{4}$.}
    \end{subfigure}
    \begin{subfigure}{0.32\textwidth}
        \centering
        \includegraphics[width=\linewidth]{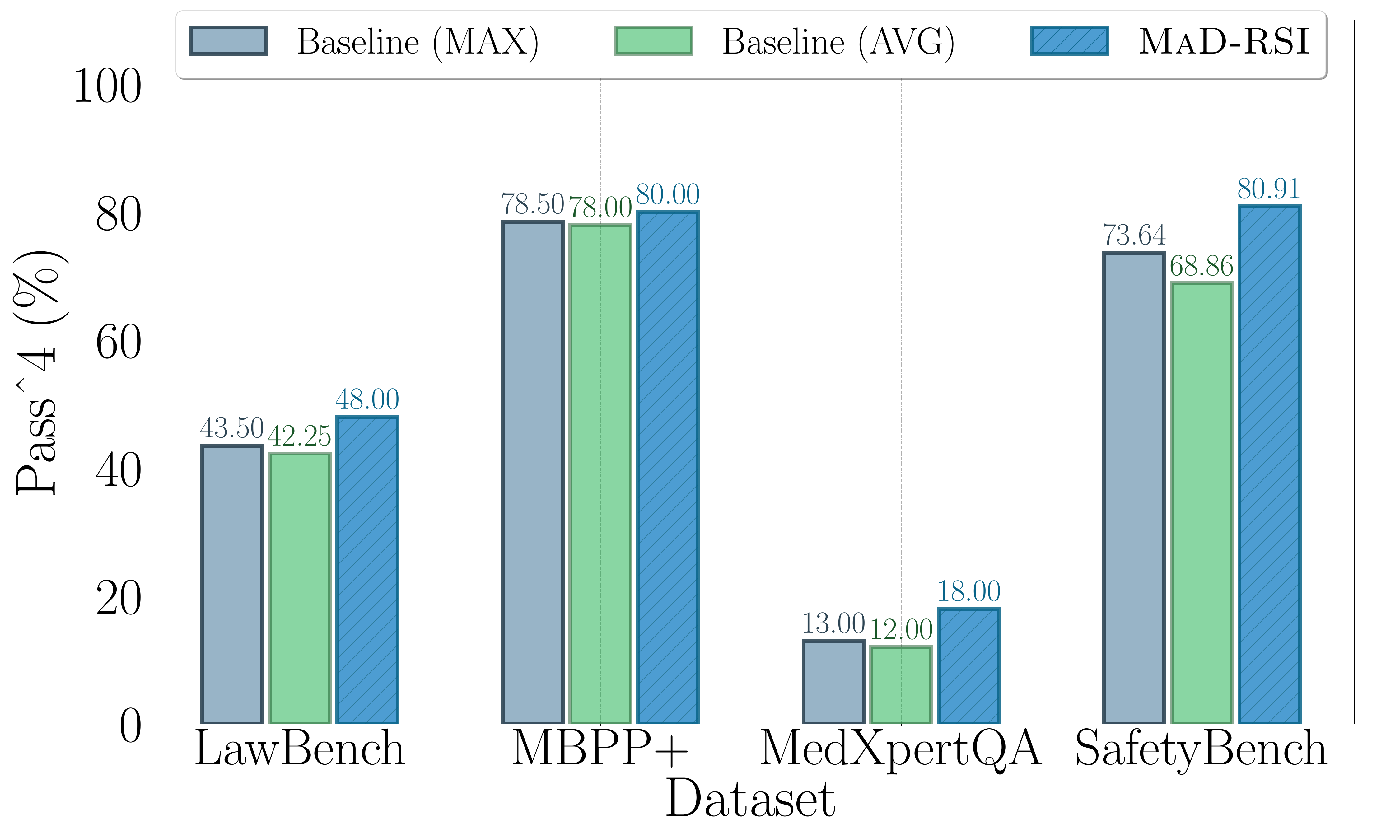}
        \caption{$\passall{4}$.}
    \end{subfigure}
    \begin{subfigure}{0.32\textwidth}
        \centering
        \includegraphics[width=\linewidth]{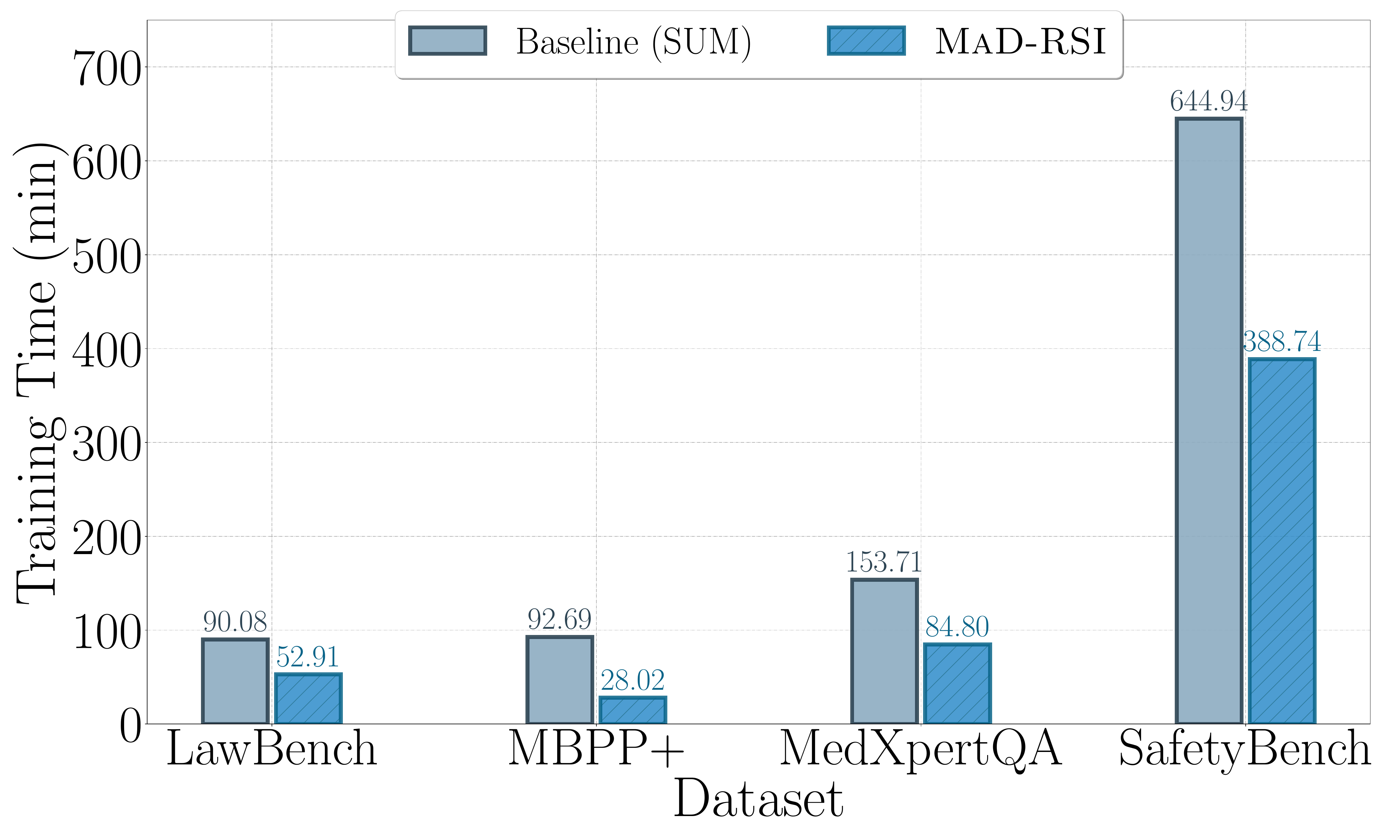}
        \caption{Training Time.}
    \end{subfigure}
    \begin{subfigure}{0.32\textwidth}
        \centering
        \includegraphics[width=\linewidth]{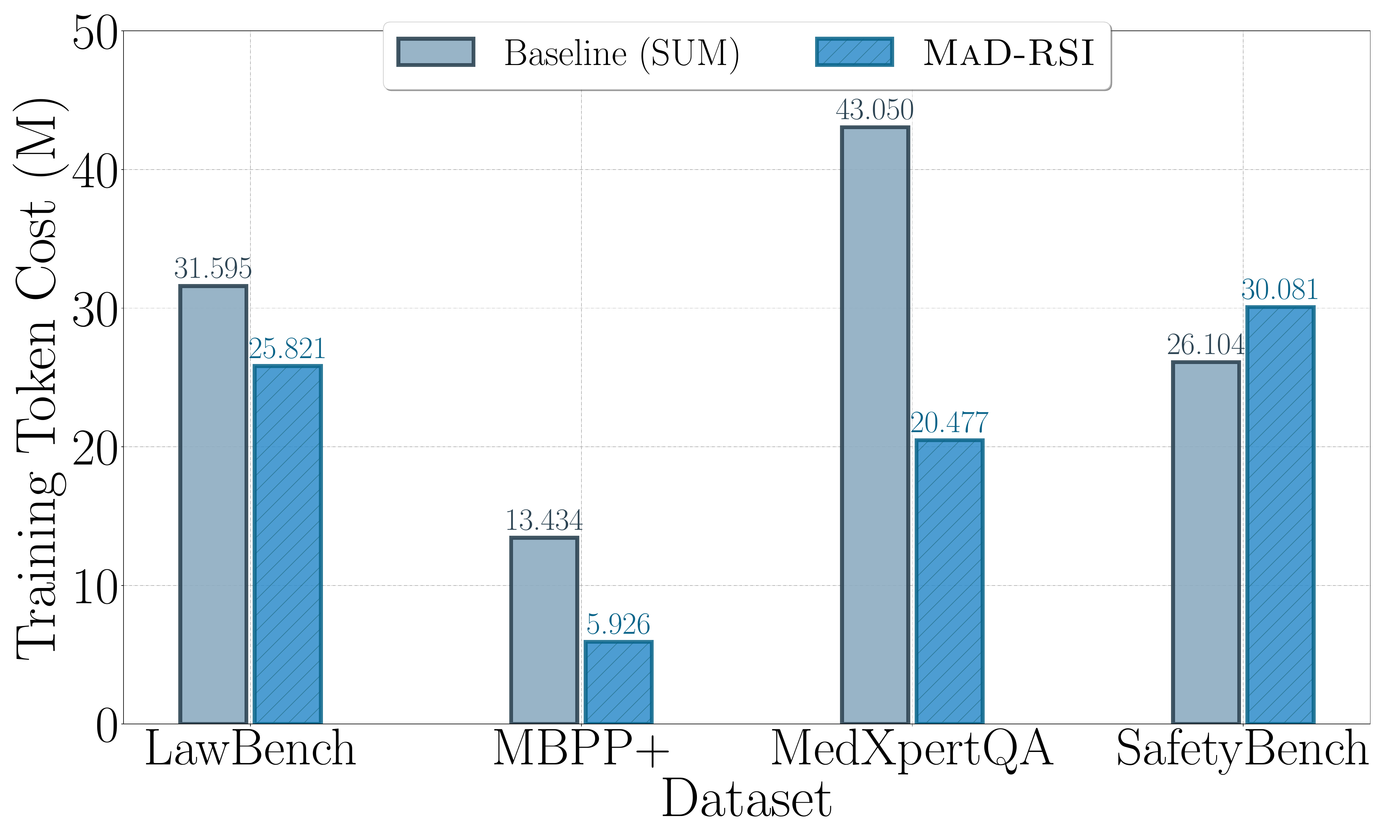}
        \caption{Training Token Cost.}
    \end{subfigure}
    \begin{subfigure}{0.32\textwidth}
        \centering
        \includegraphics[width=\linewidth]{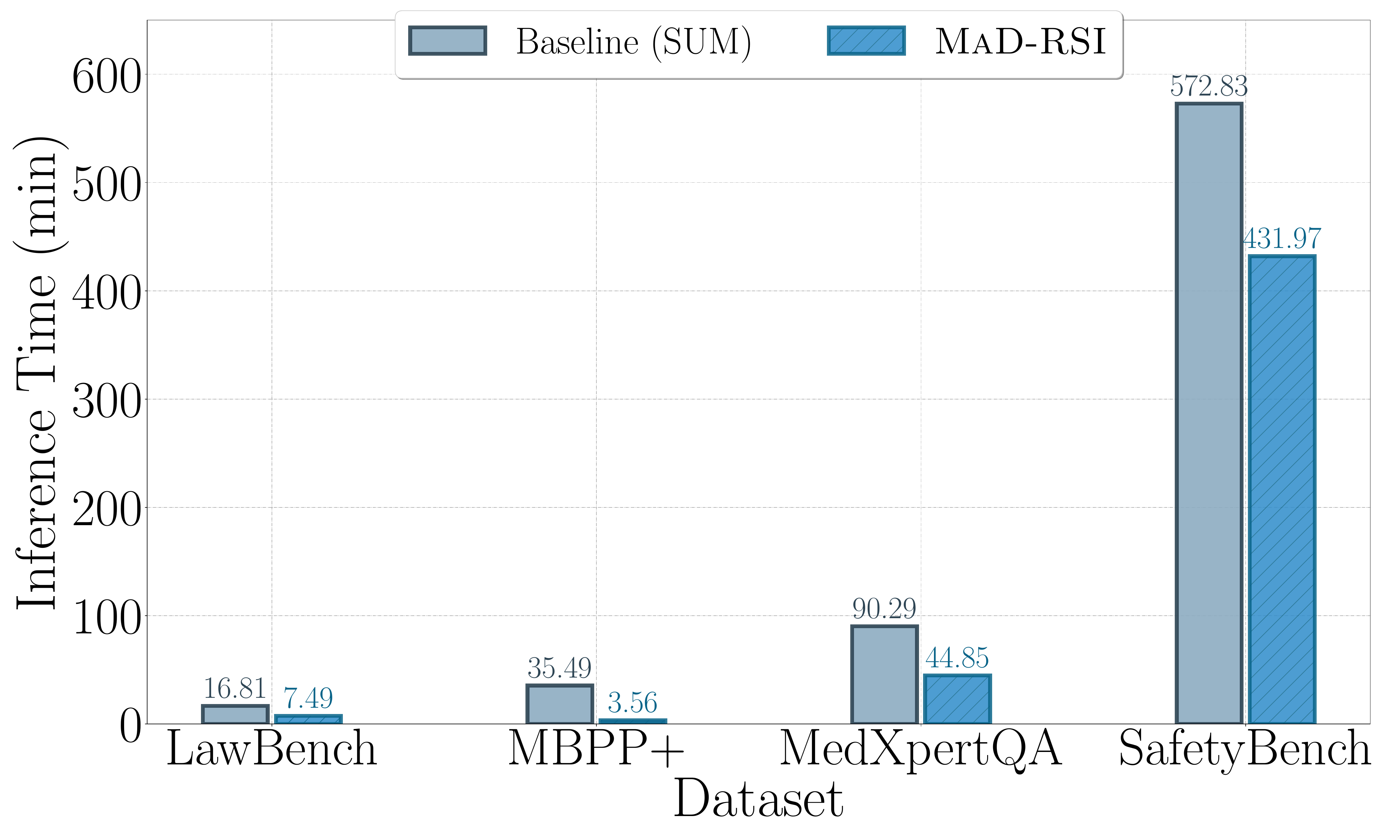}
        \caption{Inference Time.}
    \end{subfigure}
    \begin{subfigure}{0.32\textwidth}
        \centering
        \includegraphics[width=\linewidth]{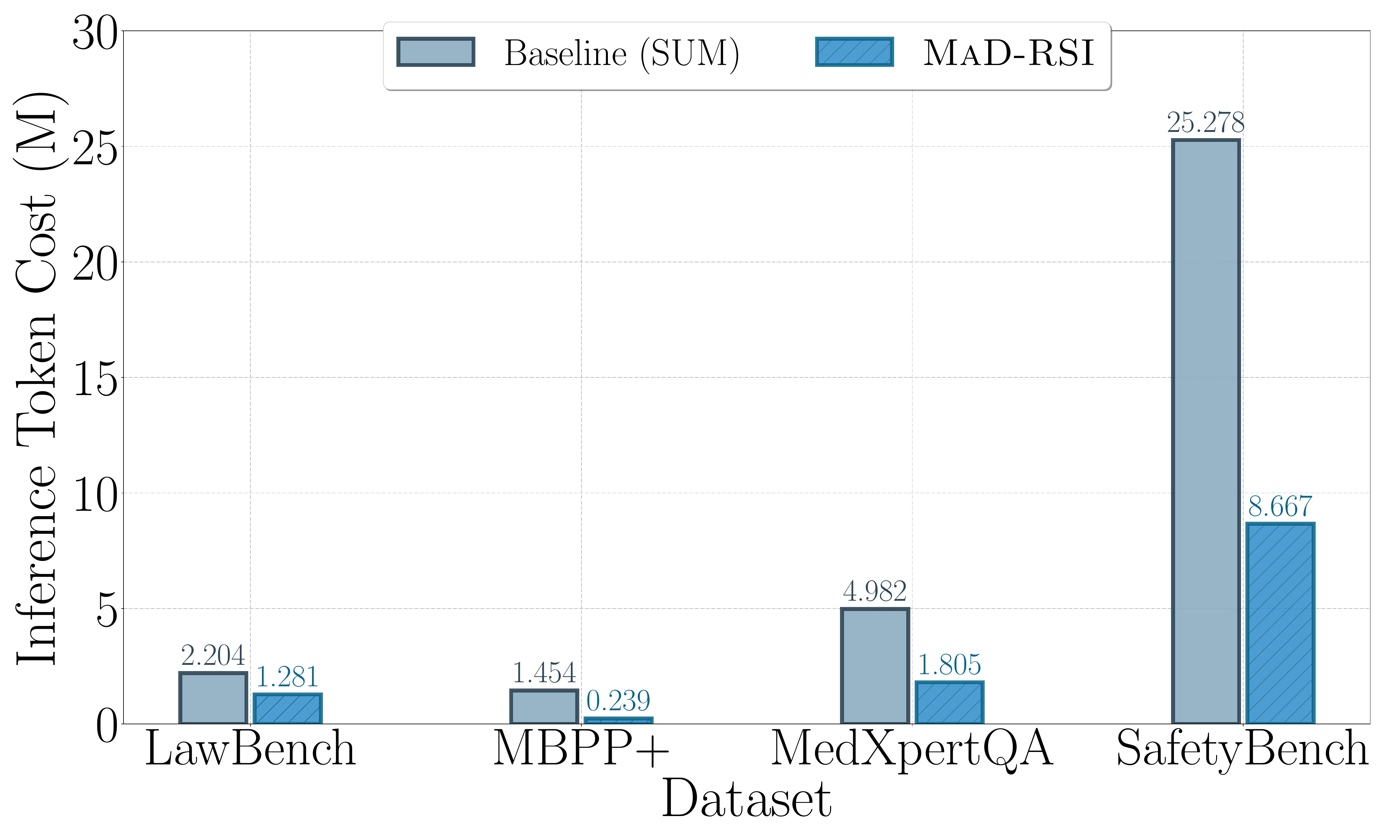}
        \caption{Inference Token Cost.}
    \end{subfigure}
    \caption{Comparison of \textsc{MaD-RSI} against four independently optimized baselines in performance and resource costs. Safety performance is measured by UA.}
    \label{fig:COM}
    \vspace{-5pt}
\end{figure*}

\section{RSI via Model-as-Data}
\label{sec:method}
\textsc{MaD-RSI} treats frozen models as sources of behavioral data for shared harness improvement \citep{zheng2026llmsimprovingllmsagentic, weng2026groupevolving}. Shared queries enable comparisons under matched execution conditions. For a fixed harness $H$ and binary outcomes, let $p_{qms}$ denote the empirical success rate for query $q$, model $m$, and sampling configuration $s$. We organize these rates into a three-dimensional response tensor, with one matrix slice per sampling configuration:
\begin{equation}
 \begin{gathered}
  \mathcal P=[p_{qms}]\in[0,1]^{N\times M\times L},\\
  \mathbf P_s=\mathcal P_{:,:,s}=
  \begin{bmatrix}
   p_{11s} & \cdots & p_{1Ms}\\
   \vdots & \ddots & \vdots\\
   p_{N1s} & \cdots & p_{NMs}
  \end{bmatrix}.
 \end{gathered}
 \label{eq:response-tensor}
\end{equation}
Each column describes model behavior across shared queries. Each row aligns responses to the same query under a fixed sampling configuration. Task evaluators score the outcomes, and each cell links to its source trajectories. Sampling configurations receive equal weights. Let $w_m \geq 0$ denote the aggregation weight assigned to
$m$, with $\sum_{m=1}^{M} w_m = 1$.
Inspired by learning for multiple target models \citep{tang2022act}, we score each query by disagreement within sampling slices:
\begin{equation}
 D_q=\frac{1}{L}\sum_s\left[
 h(\bar p_{qs})-\sum_m w_m h(p_{qms})\right].
 \label{eq:disagreement}
\end{equation}
Here $\bar p_{qs}=\sum_m w_m p_{qms}$ and $h$ is binary entropy \citep{fang-etal-2025-preemptive}. The score captures differences in empirical success rates across models under matched sampling conditions. Equal rates across models at every setting yield zero disagreement, even when individual executions vary. Computing disagreement before averaging across configurations preserves contrasts that pooling observations could hide. For binary tasks, we first reserve anchor queries, including unanimous successes and failures when available. These unanimous outcomes remain available for diagnosis despite their zero disagreement. We fill the remaining query budget for diagnosis by repeatedly adding the query with the largest combined gain in disagreement and response-pattern coverage. The coverage term favors queries that represent response patterns poorly covered by the selected set \citep{agarwal2025delift}. These patterns span models and sampling configurations. For categorical outcomes, we use the corresponding entropy contrast and a separate ranking rule. The tensor guides query selection, while trajectories supply evidence for diagnosis. For each selected query, \textsc{MaD-RSI} assembles this evidence within an input budget, preserving outcome labels and source identities. Joint diagnosis compares shared failures, model-specific failures, and successful executions. Each diagnostic claim must cite source evidence identifiers. Successful traces provide contrasts for assessing proposed edits. Disagreement alone does not establish a failure cause or a useful fix. An LLM then uses the joint diagnoses to propose revisions to the shared harness.

%% file: latex/Experiment.tex
\input{latex/tables/main_performance}

\input{latex/tables/main_six_model}

\section{Experiments}

\subsection{Experimental Setup}
\paragraph{Datasets.}
We use LawBench \citep{fei-etal-2024-lawbench}, MBPP+ \citep{liu2023mbpp}, MedXpertQA \citep{zuo2025medxpertqa}, and Agent-SafetyBench \citep{zhang2025agentsafety} (SafetyBench).

\paragraph{Models.}
For RSI, we employ DeepSeek-V4-Flash (DeepSeek-V4), DeepSeek-V3.2, Qwen3.8-Flash, and Qwen3.8-27B as the underlying target models. We assign equal weights to the four models, with $w_m = 1/4$. To assess the generalization capability of the evolved harness, we further evaluate it on Doubao-Seed-2.0-Mini (Doubao-2.0) and GLM-5.3-Flash. 

\paragraph{Evaluation and Metrics.}
During \textsc{MaD-RSI} evidence collection, temperature is the sampling parameter varied. We use $T=0.0$ and $T=1.0$, with two repetitions for each query--model--temperature combination. General-task experiments use three rounds, with two candidate slots per round. We report $\passany{M}$ to measure coverage and $\passall{M}$ to measure joint success. Main experiments use four models, while expanded evaluations include two additional unseen models. For general tasks, correctness defines the pass event. For safety tasks, we report attack safety (AS), utility under attack (UA), and clean safety (CS). We also account for resource consumption alongside performance comparisons, reporting token usage and execution time.

\paragraph{Baselines.}
We compare six configurations: the original harness, four single-executor baselines, and our \textsc{MaD-RSI}. We adopt RRSI \citep{xia2026rrsi} for general tasks and SHE \citep{qu2026she} for safety tasks. Each baseline is optimized using a single executor model and evaluated across all four models. We select RRSI because it regularizes both edit proposals and candidate selection to promote transfer to unseen benchmarks \citep{xia2026rrsi}. This comparison examines whether cross-model evidence provides additional benefits within a framework already designed for generalization.
For generalization evaluation, we select the baseline achieving the highest $\passall{4}$ in the main experiments, breaking ties using $\passany{4}$. For safety tasks, baseline selection is based on UA.

\subsection{Experimental Results}
In the main evaluation, \textsc{MaD-RSI} achieves $\passany{4}$ of 73.00\% and $\passall{4}$ of 48.00\% on LawBench, outperforming the strongest optimized baseline by 11.50 and 4.50 percentage points, respectively (see \autoref{tab:main}). Accuracy on individual models ranges from 58.50\% to 65.00\%, consistently exceeding the corresponding baseline results. On MBPP+ and MedXpertQA, improvements are more pronounced in joint success. While $\passany{4}$ matches the best baselines at 93.00\% and 60.00\%, respectively, $\passall{4}$ reaches 80.00\% and 18.00\%, yielding gains of 1.50 and 5.00 percentage points.

For SafetyBench, \textsc{MaD-RSI} achieves $\passall{4}$ of 98.93\%, 80.91\%, and 86.25\% on AS, UA, and CS, respectively, surpassing the best baseline for each metric. Notably, joint UA improves by 7.27 percentage points, despite matching the best baseline in coverage at 94.55\%. Across tasks, joint success improves more consistently than success coverage.
The advantage extends to unseen models. When evaluated with two additional executors, \textsc{MaD-RSI} outperforms the preselected baselines in $\passall{6}$ across all six dataset--metric combinations (see \autoref{tab:six-model}).

We further examine computational costs against the aggregate expenditure of four independently optimized baselines (see \autoref{fig:COM}). Baseline (MAX) and Baseline (AVG) represent the per-metric maximum and mean across these baselines, respectively. Both are computed from the same training and evaluation runs and therefore share identical costs. Importantly, the reported maxima may originate from different harnesses and do not necessarily correspond to a single configuration. Relative to the combined training-token consumption of the four baselines, \textsc{MaD-RSI} reduces usage by 18.3\%, 55.9\%, and 52.4\% on LawBench, MBPP+, and MedXpertQA, respectively, while incurring a 15.2\% increase on the safety task. The efficiency advantage therefore lies in jointly optimizing a shared harness for multiple models, rather than optimizing for a single model. Across all four datasets, \textsc{MaD-RSI} also requires fewer inference tokens and time than the four baselines combined.

%% file: latex/tables/main_performance.tex
\begin{table*}[t]
\centering
\scalebox{0.9}{
\setlength{\tabcolsep}{15pt}
\begin{tabular}{lrrrrrr}
\toprule
\multirow{2}{*}{\textbf{Scheme}} & \multicolumn{2}{c}{\textbf{LawBench}} & \multicolumn{2}{c}{\textbf{MBPP+}} & \multicolumn{2}{c}{\textbf{MedXpertQA}} \\
\cmidrule(lr){2-3}\cmidrule(lr){4-5}\cmidrule(lr){6-7}
 & $\passany{4}$ & $\passall{4}$ & $\passany{4}$ & $\passall{4}$ & $\passany{4}$ & $\passall{4}$ \\
\midrule
Original & 62.50 & 42.50 & \textbf{93.00} & 76.50 & \textbf{60.00} & 13.00 \\
RRSI / DeepSeek-V4 & 61.50 & 41.50 & \textbf{93.00} & 77.50 & \textbf{60.00} & 13.00 \\
RRSI / DeepSeek-V3.2 & 61.00 & 42.50 & 92.50 & 78.50 & 54.50 & 11.50 \\
RRSI / Qwen3.8-Flash & 61.00 & 43.50 & 91.50 & 77.50 & 55.00 & 12.00 \\
RRSI / Qwen3.8-27B & 60.50 & 41.50 & 92.00 & 78.50 & 52.50 & 11.50 \\
\textsc{MaD-RSI} & \textbf{73.00} & \textbf{48.00} & \textbf{93.00} & \textbf{80.00} & \textbf{60.00} & \textbf{18.00} \\
\midrule
\multirow{2}{*}{\textbf{Scheme}} & \multicolumn{2}{c}{\textbf{SafetyBench / AS}} & \multicolumn{2}{c}{\textbf{SafetyBench / UA}} & \multicolumn{2}{c}{\textbf{SafetyBench / CS}} \\
\cmidrule(lr){2-3}\cmidrule(lr){4-5}\cmidrule(lr){6-7}
 & $\passany{4}$ & $\passall{4}$ & $\passany{4}$ & $\passall{4}$ & $\passany{4}$ & $\passall{4}$ \\
\midrule
Original & \textbf{100.00} & 96.79 & 92.73 & 66.36 & 92.50 & 75.00 \\
SHE / DeepSeek-V4 & \textbf{100.00} & 96.43 & \textbf{94.55} & 73.64 & 92.50 & 80.00 \\
SHE / DeepSeek-V3.2 & \textbf{100.00} & 96.43 & 93.64 & 62.73 & \textbf{97.50} & 75.00 \\
SHE / Qwen3.8-Flash & \textbf{100.00} & 96.79 & 93.64 & 66.36 & \textbf{97.50} & 80.00 \\
SHE / Qwen3.8-27B & \textbf{100.00} & 97.14 & 92.73 & 72.73 & 95.00 & 75.00 \\
\textsc{MaD-RSI} & \textbf{100.00} & \textbf{98.93} & \textbf{94.55} & \textbf{80.91} & \textbf{97.50} & \textbf{86.25} \\
\bottomrule
\end{tabular}
}
\caption{Four-model evaluation across general and safety tasks.}
\label{tab:main}
\end{table*}

%% file: latex/tables/main_six_model.tex
\begin{table}[t]
\centering
\scalebox{0.65}{
\setlength{\tabcolsep}{5pt}
\begin{tabular}{llrrrr}
\toprule
\multirow{2}{*}{\textbf{Dataset}} & \multirow{2}{*}{\textbf{Metric}} & \multicolumn{2}{c}{$\passany{6}$} & \multicolumn{2}{c}{$\passall{6}$} \\
\cmidrule(lr){3-4}\cmidrule(lr){5-6}
 &  & \textbf{Baseline} & \textbf{\textsc{MaD-RSI}} & \textbf{Baseline} & \textbf{\textsc{MaD-RSI}} \\
\midrule
LawBench & \multirow{3}{*}{Accuracy} & 61.00 & \textbf{73.00} & 36.00 & \textbf{38.50} \\
MBPP+ &  & \textbf{94.00} & \textbf{94.00} & 74.50 & \textbf{77.50} \\
MedXpertQA &  & \textbf{67.00} & \textbf{67.00} & 8.50 & \textbf{13.00} \\
\midrule
\multirow{3}{*}{SafetyBench} & AS & \textbf{100.00} & \textbf{100.00} & 94.64 & \textbf{97.50} \\
 & UA & \textbf{96.36} & 95.45 & 66.36 & \textbf{76.36} \\
 & CS & \textbf{97.50} & \textbf{97.50} & 70.00 & \textbf{76.25} \\
\bottomrule
\end{tabular}
}
\caption{Expanded six-model evaluation. The population contains four development executors plus Doubao-2.0 and GLM-5.3-Flash.}
\label{tab:six-model}
\end{table}

%% file: latex/Related_Work.tex
\section{Related Work}

Existing methods for RSI of agent harnesses largely follow a common paradigm: optimizing a harness using a single execution model and subsequently evaluating its transfer across models \citep{hu2025automated, zhang2026darwin, lee2026metaharness, xia2026rrsi}. However, improvements achieved during optimization do not necessarily transfer to other models. Previous empirical findings suggest that harness modifications tailored to the failure modes of a source model may provide limited benefits to other models or even degrade their performance \citep{qu2026she, wang2026promptbridge, rakotonirina2023can}. This observation motivates us to incorporate behavioral differences across models directly into the training evidence. Unlike existing approaches, \textsc{MaD-RSI} compares the behaviors of different frozen execution models during evidence construction. Specifically, it aligns responses to the same query across models and sampling configurations, leverages these aligned observations for query selection and joint diagnosis, and uses the resulting evidence to guide the optimization of a shared harness.